\documentclass{article}

\usepackage{microtype}
\usepackage{graphicx}

\usepackage{subfigure}
\usepackage{booktabs} 

\usepackage{hyperref}
\usepackage{xcolor}

\usepackage[accepted]{icml2025}
\usepackage{pifont} 
\usepackage{multirow}
\usepackage{xcolor}
\usepackage{stmaryrd}
\usepackage{colortbl}
\usepackage{booktabs}
\definecolor{mygray}{gray}{.9}
\usepackage{amsmath}
\usepackage{amssymb}
\usepackage{mathtools}
\usepackage{amsthm}

\usepackage[capitalize,noabbrev]{cleveref}

\theoremstyle{plain}

\theoremstyle{definition}

\theoremstyle{remark}

\usepackage[textsize=tiny]{todonotes}

\icmltitlerunning{Motion-Prior Conditional Diffusion Model}

\begin{document}

\twocolumn[
\icmltitle{Long-Term TalkingFace Generation \\ 
via Motion-Prior Conditional Diffusion Model}




\begin{icmlauthorlist}
\icmlauthor{Fei Shen}{1}
\icmlauthor{Cong Wang}{2}
\icmlauthor{Junyao Gao}{3}
\icmlauthor{Qin Guo}{4}
\icmlauthor{Jisheng Dang}{5}
\icmlauthor{Jinhui Tang}{1}
\icmlauthor{Tat-Seng Chua}{6}
\end{icmlauthorlist}

\icmlaffiliation{1}{Nanjing University of Science and Technology}
\icmlaffiliation{2}{Nanjing University}
\icmlaffiliation{3}{Tongji University}
\icmlaffiliation{4}{Peking University}
\icmlaffiliation{5}{Sun Yat-sen University}
\icmlaffiliation{6}{National University of Singapore}

\icmlcorrespondingauthor{Jinhui Tang, Contact}{feishen@njust.edu.cn}

\icmlkeywords{Machine Learning, ICML}

\vskip 0.3in
]



\printAffiliationsAndNotice{}  

\begin{abstract}
Recent advances in conditional diffusion models have shown promise for generating realistic TalkingFace videos, yet challenges persist in achieving consistent head movement, synchronized facial expressions, and accurate lip synchronization over extended generations. To address these, we introduce the \textbf{M}otion-priors \textbf{C}onditional \textbf{D}iffusion \textbf{M}odel (\textbf{MCDM}), which utilizes both archived and current clip motion priors to enhance motion prediction and ensure temporal consistency. The model consists of three key elements: (1) an archived-clip motion-prior that incorporates historical frames and a reference frame to preserve identity and context; (2) a present-clip motion-prior diffusion model that captures multimodal causality for accurate predictions of head movements, lip sync, and expressions; and (3) a memory-efficient temporal attention mechanism that mitigates error accumulation by dynamically storing and updating motion features. We also release the \textbf{TalkingFace-Wild} dataset, a multilingual collection of over 200 hours of footage across 10 languages. Experimental results demonstrate the effectiveness of MCDM in maintaining identity and motion continuity for long-term TalkingFace generation. Code, models, and datasets will be publicly available.
\end{abstract}

\section{Introduction}

TalkingFace generation~\cite{tan2024flowvqtalker, peng2024synctalk, zhou2021pose, ye2024real3d, ji2021audio, tan2023emmn, kim2018deep, liang2022expressive, ye2023geneface, pumarola2018ganimation, vougioukas2020realistic} aims to create realistic and expressive videos from a reference face and audio, with applications in virtual avatars, gaming, and filmmaking. However, the complexity of facial movements, including head, lip, and expression motions, presents challenges, along with the need to maintain identity consistency across extended sequences.

Early methods~\cite{vougioukas2020realistic, wang2021one, hong2022depth,chan2022efficient, guo2024liveportrait} use GANs~\cite{goodfellow2014generative, mirza2014conditional} to synthesize facial motions onto a reference image through a two-step process: decoupling motion features from audio and mapping them onto intermediate representations like facial landmarks~\cite{yang2023effective}, 3DMM~\cite{sun2023vividtalk}, or HeadNeRF~\cite{hong2022headnerf}. Despite their promise, GAN-based methods suffer from training instability and inaccuracies in motion extraction, often leading to artifacts like blurriness and flickering that compromise video realism.
Recent diffusion models~\cite{wei2024aniportrait, stypulkowski2024diffused, tian2024emo, guo2024liveportrait,zheng2024memo, jiang2024loopy} have improved TalkingFace generation by enhancing video realism through multi-step denoising that preserves conditional input information. These methods typically use a Reference UNet~\cite{hu2024animate} to encode identity features and integrate audio via cross-attention. However, reliance on static audio features and weak correlations between audio and motion complicate the decoupling of identity and motion cues, often resulting in artifacts like motion distortion and flickering, especially in long-term generation.

While some methods~\cite{wang2024v,ma2024follow,yang2024megactor} improve long-term stability by introducing motion constraints like facial landmarks and emotion tags, these constraints often overly bind poses to the reference image, limiting expression diversity. 
Models trained with driven landmark fail to learn natural audio-driven motion patterns, reducing audio-visual synergy. Additionally, static emotion tags cannot capture dynamic shifts, leading to rigid, inauthentic animations over extended sequences. 
Besides, some approaches~\cite{xu2024hallo, chen2024echomimic} inject brief motion reference frames, usually fewer than five over 0.2 seconds, which is insufficient to establish coherent motion, resulting in random, less dynamic movements.

In this paper, we propose the \textbf{M}otion-priors \textbf{C}onditional \textbf{D}iffusion \textbf{M}odel (\textbf{MCDM}) to address the challenges in achieving long-term consistency in TalkingFace generation. The MCDM comprises three key modules: the archived-clip motion-prior, the present-clip motion-prior diffusion model, and a memory-efficient temporal attention mechanism.
Unlike conventional reference UNet-based identity learning, the archived-clip motion-prior introduces historical frames along with a reference frame via frame-aligned attention , enhancing identity representation and creating a cohesive facial context over extended sequences. 
Then, the present-clip motion-prior diffusion model leverages multimodal causality and temporal interactions to effectively decouple and predict motion states, including head, lip, and expression movements, ensuring a clear separation between identity and motion features and promoting temporal consistency across frames.
To support long-term stability, we devise a memory-efficient temporal attention that dynamically stores and updates historical motion features, integrating them with current motion cues via a memory update mechanism. This structure reduces error accumulation often observed in diffusion-based long-term TalkingFace generation, enabling more stable and consistent outputs.
Additionally, we present the \textbf{TalkingFace-Wild} dataset, a high-quality, multilingual video dataset with over 200 hours of footage in 10 languages, offering a valuable resource for further research in TalkingFace generation.
Our main contributions are summarized as follows:
\begin{itemize}
  \item[$\bullet$] We propose MCDM to enhance robust identity consistency and support temporal consistency in long-term TalkingFace generation.
\item[$\bullet$] We develop the archived-clip motion-prior module to enhance identity representation and construct a comprehensive facial context from historical frames.
 \item[$\bullet$] We devise the present-clip motion-prior diffusion model to decouple current identity and motion features via multimodal causality and temporal interactions.
 \item[$\bullet$] We present a memory-efficient temporal attention to dynamically update and integrate historical motion features with current ones, reducing error accumulation.
  \item[$\bullet$] We release the TalkingFace-Wild dataset, covering 10 languages and over 200 hours of video for advancing TalkingFace research.
\end{itemize}

\begin{figure*}[t]
\includegraphics[width=\linewidth]{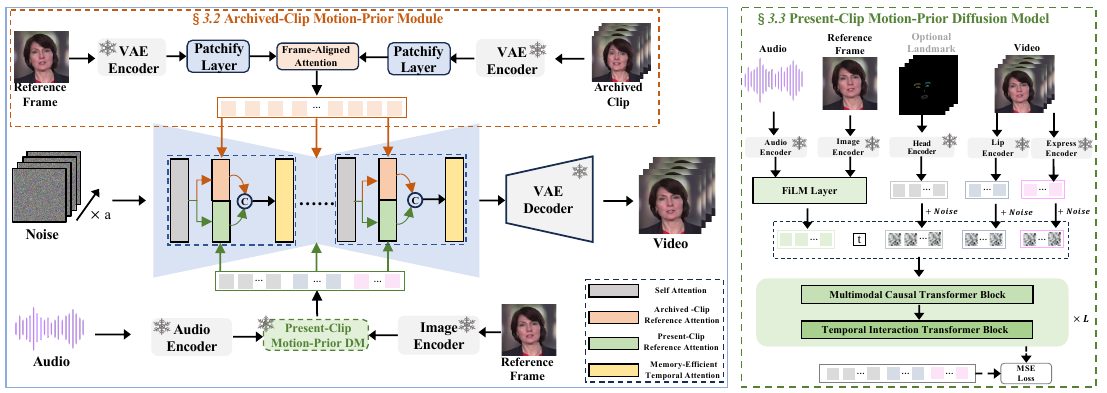}
     \vspace{-0.5cm}
     \caption{\textbf{Our MCDM architecture.}
     On the upper, the archived-clip motion-prior leverages frame-aligned attention with archived-clip, enhancing identity coherence over extended sequences. On the right, the present-clip motion-prior diffusion model uses multimodal causality and temporal interactions to decouple and predict motion states, covering head, lip, and expression movements while maintaining a clear separation of identity and motion features.
     }
 \vspace{-0.4cm}
     \label{fig:framework}
 \end{figure*}

\section{Related Work}

\textbf{GAN-Based Methods.}  
GAN-based approaches~\cite{kim2018deep, zhou2020makelttalk, pumarola2018ganimation, vougioukas2020realistic, zhang2023sadtalker, wang2021one, hong2022depth, chan2022efficient, guo2024liveportrait} for TalkingFace generation extract motion features from audio or visual inputs and map them to intermediate representations such as facial landmarks~\cite{yang2023effective}, 3DMM~\cite{sun2023vividtalk}, or HeadNeRF~\cite{hong2022headnerf}. MakeItTalk~\cite{zhou2020makelttalk} employs LSTMs to predict landmarks from audio, followed by a warp-based GAN for video synthesis. GANimation~\cite{pumarola2018ganimation} models facial motion via continuous manifolds, enhancing expression dynamics. SadTalker~\cite{zhang2023sadtalker} integrates ExpNet and PoseVAE to refine motion representations within the FaceVid2Vid~\cite{wang2021one} framework. DaGAN~\cite{hong2022depth} introduces self-supervised geometric learning to capture dense 3D motion fields. While effective, GAN-based methods suffer from adversarial training instability and motion inaccuracies, often resulting in artifacts that degrade realism.

\textbf{Diffusion-Based Methods.}  
Diffusion models~\cite{rombach2022high} have gained traction in TalkingFace generation, producing high-quality, diverse outputs. AniPortrait~\cite{wei2024aniportrait} maps audio to 3D facial structures, generating temporally coherent videos with expressive detail. MegActor-$\Sigma$~\cite{wang2024v} synchronizes lip movements, expressions, and head poses using a reference UNet~\cite{hu2024animate} and facial loss functions to enhance fidelity. Hallo~\cite{xu2024hallo} and EchoMimic~\cite{chen2024echomimic} leverage limited motion reference frames to improve expression diversity and pose alignment. However, reliance on short-term frame histories ($2$-$4$ frames) compromises long-term motion consistency, while increased frame dependencies escalate computational costs. Additionally, static audio features and restricted references fail to capture natural motion variations, leading to artifacts such as motion distortion and rigid expressions in extended sequences.

Unlike prior work, our approach introduces motion priors from both archived and present clips to enhance long-term motion prediction and identity consistency. By leveraging historical frames and memory-efficient temporal attention, MCDM improves motion continuity while maintaining realism in TalkingFace generation.

\section{Method}

\noindent\textbf{Task Definition.}
Given a reference image, audio, and optional facial landmarks, TalkingFace generation aims to produce temporally coherent and realistic videos.
The key challenges include maintaining consistent identity over time, achieving natural head movements, and ensuring expressive and precise lip alignment with audio cues.
However, existing methods often encounter limitations such as error accumulation, inconsistent identity preservation, suboptimal audio-lip synchronization, and rigid expressions.

\subsection{Overall Framework}
To address the above challenges, we introduce MCDM, a framework centered on a denoising UNet resembling Stable Diffusion v1.5 (SD v1.5)~\footnote{https://huggingface.co/runwayml/stable-diffusion-v1-5}, tailored to denoise multi-frame noisy latent inputs under conditional guidance.
As illustrated in Figure~\ref{fig:framework}, unlike standard UNet architectures, each Transformer block in MCDM incorporates four attention layers.
The first layer, a self-attention, mirrors that in SD v1.5. The second and third layers are parallel cross attention (spatial-wise), designed for distinct interactions: the archived-clip reference attention layer, which integrates motion priors from archived clip encoded by the archived-clip motion-prior module (Section~\ref{amp}), and the present-clip reference attention, which engages with present clip priors from the present-clip motion-prior diffusion model (Section~\ref{pmp}). 
The fourth layer, the memory-efficient temporal attention (Section~\ref{lta}), is a temporal-wise self attention that dynamically updates and merges archived motion features with current motion features, effectively mitigating error accumulation.

\subsection{Archived-Clip Motion-Prior Module}\label{amp}
\noindent\textbf{Motivation.} Existing methods typically use the past $2-4$ frames to guide the denoising network for generating temporally consistent videos. However, this limited history frame is insufficient for maintaining long-term consistency, and incorporating more frames exponentially increases computational demand, making it impractical for real-world applications. To overcome these limitations, we propose an archived-clip motion prior that integrates long-term historical frames and a reference frame into the denoising UNet via conditional frame-aligned attention, enhancing identity representation and establishing motion context.

\noindent\textbf{Architecture.} As illustrated in Figure~\ref{fig:framework}, the archived-clip motion-prior consists of two frozen VAE encoders, two learnable patchify layers, and a frame-aligned attention mechanism.
Given a reference frame \( X_{\text{ref}} \in \mathbb{R}^{b \times 1 \times c \times h \times w} \) and a archived clip \( X_{\text{arch}} \in \mathbb{R}^{b \times a \times c \times h \times w} \), where \( b \), \( c \), \( h \), \( w \), and \( a \) represent the batch size, channels, height, width, and the number of archived frames, respectively.
First, the frozen VAE encoder extracts latent features from both the reference and archived frames, resulting in \( f_x \in \mathbb{R}^{b \times 1 \times 4 \times \frac{h}{8} \times \frac{w}{8}} \) and \( f_a \in \mathbb{R}^{b \times a \times 4 \times \frac{h}{8} \times \frac{w}{8}} \), respectively.
Next, the learnable patchify layers, consisting of 2D convolutions followed by flattening operations, transform these latent features into tokens, yielding \( F_x \in \mathbb{R}^{b \times 1 \times m \times d} \) and \( F_a \in \mathbb{R}^{b \times a \times m \times d} \), where \( m \) and \( d \) denote the token length and embedding dimension.

In the frame-aligned attention, we adopt a frame-wise computation approach to improve efficiency and adaptability for long temporal sequences. 
For each archived frame \( i \in [1, a] \), the Key \( K_i \) is derived from the reference tokens \( F_x \), while the Value \( V_i \) is derived from the tokens of the corresponding archived frame \( F_{a}^{i} \):
\begin{equation}
\vspace{-0.05cm}
K_i = F_x \mathbf{W}_K, \quad V_i = F_{a_i} \mathbf{W}_V,
 \vspace{-0.05cm}
\end{equation}
where \( \mathbf{W}_K \in \mathbb{R}^{d \times d} \) and \( \mathbf{W}_V \in \mathbb{R}^{d \times d} \) are learnable projection matrices for the Key and Value.
The attention for each frame \( i \) is then computed as:
\begin{equation}
\vspace{-0.05cm}
\text{Attention}(Q, K_i, V_i) = \text{Softmax}\left(\frac{Q K_i^\top}{\sqrt{d}}\right) V_i ,
\vspace{-0.05cm}
\end{equation}
where \( Q \in \mathbb{R}^{n \times d} \) represents a learnable query tokens, with \( n \) denoting the number of queries.
Aggregating the outputs across all frames yields the final output \( F_{ac} \in \mathbb{R}^{b \times a \times n \times d} \), where each frame's attended tokens reflect both the static reference and dynamic temporal information.

\subsection{Present-Clip Motion-Prior Diffusion Model}\label{pmp}
\noindent\textbf{Motivation.} 
Motion information is typically driven either by landmark signals from a driving video or directly by audio cues. The landmark-driven approach guides reference image movements but limits the natural diversity of head motions and expressions. In contrast, audio-driven methods rely solely on audio cues, often lacking sufficient guidance for realistic head movement. To address these limitations, we propose the present-clip motion-prior diffusion model, which first predicts motion states, including head, lip, and expressions motions, rather than directly generating TalkingFace videos.

\noindent\textbf{Architecture.} 
We aim to predict motion in head, lip, and expressions lip movements, conditioned on audio and image tokens. 
As shown in Figure~\ref{fig:framework} (right), we begin by extracting feature tokens from the audio encoder, image encoder, head encoder, lip encoder, and express encoder.

\noindent\textbf{Audio Encoder:} Audio sequence tokens are extracted from the input audio via a frozen Wav2Vec model~\cite{baevski2020wav2vec}.

\noindent\textbf{Image Encoder:} Image tokens are extracted from the reference frame using a frozen CLIP~\cite{radford2021learning} and are replicated along the temporal dimension to align with audio features.

\noindent\textbf{Head Encoder:} Head tokens are extracted from reference landmark video through a frozen Landmark Guider~\footnote{https://github.com/MooreThreads/Moore-AnimateAnyone}; notably, these tokens are optional, allowing simulation of conditions with or without reference video guidance.

\noindent\textbf{Lip and Express Encoders:} Lip and expression tokens are extracted from the target video using a custom-trained encoder. Details of the lip and express encoders are provided in the supplementary material.

We then pass the audio and image tokens through a feature-wise linear modulation (FiLM) layer~\cite{perez2018film} to adaptively learn multimodal correlation tokens. These tokens, along with the timestep $t$, and noise-added tokens for head, lip, and expression movements, are prepended to the input sequence. This composite input is fed into an $L$-layer structure consisting of a multimodal causal transformer block~\cite{peebles2023scalable} and a temporal interaction transformer block~\cite{hu2024animate}, with added noise in facial motion tokens acting as the supervision.
The training loss \( L_{\text{prior}} \) for the present-clip motion-prior diffusion model \( \epsilon_\theta \) is defined as:
\begin{equation}
L_{\text{prior}} = \mathbb{E}_{t, F_{p}, z_{t}, \epsilon,} \left\| \epsilon - \epsilon_\theta \left( z_t, t, F_p \right) \right\|^2.
\end{equation}
Without landmark guidance, \( F_{p} \) represent multimodal interaction tokens from audio and the reference frame. \( z_{t} \) represent noise-added tokens for head, lip, and expression movements at timestep \( t \). With landmark guidance, \( F_{p} \) additionally include landmark tokens.  \( z_{t} \) represent noise-added lip and expression tokens.
This design allows flexible conditioning, incorporating landmark guidance when available, while effectively leveraging multimodal interactions for accurate motion state predictions.

\begin{figure}[t]
\includegraphics[width=\linewidth]{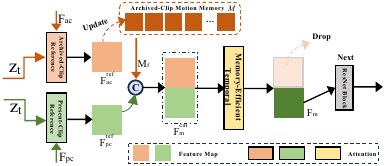}
     \vspace{-15pt}
     \caption{\textbf{The overview of memory-efficient temporal attention.} It can dynamically update and integrate historical motion features with current ones.
     }
 \vspace{-0.4cm}
     \label{fig:motion}
 \end{figure}

\subsection{Memory-Efficient Temporal Attention}\label{lta}
\noindent\textbf{Motivation.} 
For long-term TalkingFace generation, current methods primarily adopt either fully or semi-autoregressive strategies: the former generates one frame per iteration, while the latter produces a fixed-length clip. However, due to GPU memory limitations, relying on a restricted frame history for extrapolation often results in error accumulation, as limited prior motion information undermines consistency over extended sequences.
Therefore, we propose a memory-efficient temporal attention to dynamically update and integrate historical motion features with current ones, reducing error accumulation.

\noindent\textbf{Architecture.} 
AnimateDiff~\cite{guo2023animatediff} demonstrates that the temporal layer in self-attention ensures smooth temporal continuity and consistency of appearance details across frames. We replace traditional self-attention with fast attention~\cite{choromanski2020rethinking} in the temporal layer to enhance temporal continuity and manage memory efficiently, allowing the accumulation of extensive historical motion information for consistent long-sequence generation. 
As shown in Figure~\ref{fig:motion}, let $F_{ac}$ and $F_{pc}$ denote the output features of the archived-clip and present-clip motion-prior modules, respectively, and let $Z_t$ represent the noisy latent feature at time step $t$. These features undergo reference attention, yielding refined representations $F_{ac}^{\text{ref}}$ and $F_{pc}^{\text{ref}}$, which capture spatial-domain motion characteristics. $F_{ac}^{\text{ref}}$ is then input into the motion memory update mechanism, which aggregates motion across frames, producing the update feature $M_f$. The memory update mechanism is defined as follows,

\noindent\textbf{(1) Initialization}: At the first frame, the memory $M_1$ is initialized with $F_{ac}^{\text{ref}}$ since no prior motion information exists:
\begin{equation}
\vspace{-0.1cm}
  M_1 = F_{ac}^{\text{ref}} .
  \vspace{-0.1cm}
\end{equation}
\noindent\textbf{(2) Memory Update}: For each frame \( f \), the memory \( M_f \) is updated by combining the current feature \( F_{ac}^{\text{ref}} \) with the previous memory \( M_{f-1} \) as:
\begin{equation}\label{alpha}
\vspace{-0.1cm}
  M_f = \alpha M_{f-1} + (1 - \alpha) F_{ac}^{\text{ref}},
  \vspace{-0.1cm}
\end{equation}
where $\alpha \in [0, 1]$ controls the balance between past and current frames. This fixed memory update mechanism avoids storage bottlenecks of historical information.
We then concatenate \( F_{pc}^{\text{ref}} \) with \( M_f \) along the temporal dimension, creating \( F_{m}^{\text{cat}} \), which integrates past and current motion. \( F_{m}^{\text{cat}} \) is processed through Fast Attention along the temporal axis to capture dependencies across frames, with the lower half of the resulting feature map used as the output $F_m$.

\subsection{Training and Inference}\label{mcdm}
\noindent\textbf{Training.}
Our training process is divided into three stages, each with specific learning objectives.
Each stage is supervised using standard MSE loss~\cite{rombach2022high}.

\noindent\textbf{Stage1.} The archived-clip motion-prior is trained to enhance identity representation and establish a robust facial motion context across extended sequences. The present-clip reference attention and memory-efficient temporal attention modules remain frozen during this stage.

\noindent\textbf{Stage2.} The present-clip motion-prior diffusion model is trained to predict the motion states of facial expressions, lip, and head movements. 
To simulate scenarios without a driving video, we randomly drop the entire landmark clip.

\noindent\textbf{Stage3.} The full motion-priors conditional diffusion model is trained for generating stable and consistent long-term TalkingFace videos. 
Only the present-clip reference and memory-efficient temporal attentions are trained. 

\noindent\textbf{Inference.}
The present-clip motion-prior diffusion model first predicts distinct motion tokens based on the given conditions (either with or without landmark guidance). Landmarks are not used by default unless specified.
Subsequently, MCDM utilizes these motion tokens, alongside a single reference image and audio input, to generate the video sequence.
For the initial archived clip, we initialize it using the reference image and then progressively update the motion memory to ensure temporal consistency.

\section{Experiments}

\subsection{Experimental Settings}
\noindent\textbf{Datasets.}
The HDTF dataset~\cite{zhang2021flow} comprises 410 videos with over 10,000 unique speech sentences, varied head poses, and movement patterns. 
Following prior work~\cite{chen2024echomimic, tian2024emo, xu2024hallo}, we split HDTF into training and testing sets with a 9:1 ratio. 
The CelebV-HQ dataset~\cite{zhu2022celebv} includes 35,666 clips (3–20 seconds each) across 15,653 identities, totaling roughly 65 hours.
Both datasets present quality issues, such as audio-lip misalignment, facial occlusions, small facial regions, and low resolution. 
To mitigate these, we developed a custom data processing pipeline for high-quality TalkingFace data, detailed in the following subsection.

Additionally, mostly methods~\cite{wang2024v, xu2024hallo, jiang2024loopy} employ proprietary datasets for supplementary training and testing. 
Similarly, we sourced a variety of TalkingFace videos from YouTube using targeted keyword queries (e.g., “nationality,” “interview,” “dialogue”) across different languages and contexts. 
From Table~\ref{tab:statistic}, we collect a new high-quality dataset, \textbf{TalkingFace-Wild}, covering 10 languages and totaling over 200 hours after processing through our data pipeline. 
To assess the generalization capability of models, we also constructed an open-set test collection of 20 diverse portrait images and 20 audio clips.

\begin{table}[t]

\vspace{-0.2cm}
\footnotesize
\centering
    \resizebox{0.50\linewidth}{!}{
    \begin{tabular}{c}
    \includegraphics[width=1\linewidth]{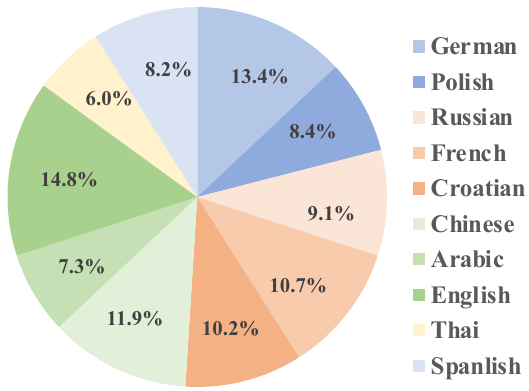}
    \end{tabular}}
    \,
    \resizebox{0.46\linewidth}{!}{
    \begin{tabular}{ll @{\quad}cc}
    \toprule
    \multicolumn{2}{c}{Statistics of our TalkingFace-Wild dataset}\\
    \cmidrule{1-2}
    Num. of languages&10\\
    Num. of identities&3,452\\
    Num. of video clips&31.3k\\
    Total hours&241.6 h\\
    Avg. duration&27.8 s\\
    \bottomrule
    \end{tabular}
    }
    \vspace{-0.2cm}
        \caption{\textbf{Statistics of our TalkingFace-Wild dataset.} We release a TalkingFace dataset that is well-balanced across 10 languages.}
\label{tab:statistic}
\vspace{-0.4cm}
\end{table}

\begin{table*}[t]

\vspace{-.3cm}
    \centering
    \footnotesize
    \setlength{\tabcolsep}{3.2 pt}
    \begin{tabular}{l|cccccc|cccccc}
      \toprule  
        \multirow{2}*{\begin{tabular}{l}\cellcolor{mygray}\textbf{Method}\\
                \end{tabular}}
        &\multicolumn{6}{c|}{HDTF}
        &\multicolumn{6}{c}{CelebV-HQ}\\
        & \cellcolor{mygray}{\textbf{FID$\downarrow$}} & \cellcolor{mygray}{\textbf{FVD$\downarrow$}}  & \cellcolor{mygray}{\textbf{Sync-C$\uparrow$}} & \cellcolor{mygray}{\textbf{Sync-D$\downarrow$}}& \cellcolor{mygray}{\textbf{SSIM$\uparrow$}} & \cellcolor{mygray}{\textbf{E-FID$\downarrow$}}
        & \cellcolor{mygray}{\textbf{FID$\downarrow$}} & \cellcolor{mygray}{\textbf{FVD$\downarrow$}}  & \cellcolor{mygray}{\textbf{Sync-C$\uparrow$}} & \cellcolor{mygray}{\textbf{Sync-D$\downarrow$}}& \cellcolor{mygray}{\textbf{SSIM$\uparrow$}} & \cellcolor{mygray}{\textbf{E-FID$\downarrow$}}\\
 \midrule
    Audio2Head &76.08 &1417.65 &3.16 &17.62 &0.572 &3.81  &127.30 &1882.64 &1.96 &17.36 &0.391 &8.42\\
    V-Express &57.14 &1152.29 &5.05 &11.68 &0.706 &1.83    &98.07 &1465.26 &3.71 &13.41 &0.514 &5.18\\
    AniPortrait& 54.81 &1072.63 &5.40 &11.39 &0.727 &1.95 &94.25 &1260.74 &3.98 &12.88 & 0.536 &4.91\\
    SadTalker &52.77 &956.24 &5.73 &10.65 &0.736 &1.87 &88.22 & 1055.49 &4.05 &11.20 &0.565 & 4.66\\ 
    Hallo&37.29 &616.04 &6.33 &8.64 &0.774 & 1.67 &72.46 & 907.60 &6.48  &8.61 &0.620 &2.93 \\
    EchoMimic &31.44 &595.17 &6.96 &8.59 &0.782 & 1.64 &71.47 & 893.28 &6.70 & 8.45 & 0.637 & 2.81\\
    {MegActor-}$\Sigma$ &31.37 &586.10 &6.87 &8.55 &0.778 &1.62 &70.82 &875.21 &6.77 &8.32 &0.634 &2.74\\
    \textbf{MCDM (Ours)} &\textbf{26.45} &\textbf{543.28} &\textbf{7.49} &\textbf{8.04} &\textbf{0.824} &\textbf{1.51} &\textbf{67.29} &\textbf{784.53} &\textbf{7.25} &\textbf{7.84} &\textbf{0.662} &\textbf{2.31}\\
    \bottomrule
    \end{tabular}
\vspace{-.3cm}
\caption{\textbf{Quantitative comparisons on HDTF and CelebV-HQ.} MCDM achieves the top results across all metrics, with best in \textbf{bold}.
}\label{tab:sota}
\vspace{-.3cm}
\end{table*}

\noindent\textbf{Data Processing.}\label{DPP}
First, we detect scene transitions in raw videos using PySceneDetect\footnote{https://github.com/Breakthrough/PySceneDetect} and trim each clip to a maximum duration of 30 seconds. Next, we apply face detection~\cite{guo2021sample} to exclude videos lacking complete faces or containing multiple faces, using the bounding boxes to extract talking head regions. Third, an image quality assessment model~\cite{su2020blindly} filters out low-quality and low-resolution clips. Fourth, SyncNet~\cite{prajwal2020lip} assesses audio-lip synchronization, discarding clips with misaligned audio. Finally, we manually inspect a subset to verify audio-lip synchronization and overall video quality, ensuring precise filtering.
In addition, to ensure a fair comparison, we report results trained independently on each of the previously mentioned datasets.

\noindent\textbf{Metrics.}
We utilize a comprehensive set of metrics to assess the quality of generated videos and audio-lip synchronization. Fréchet Inception Distance (FID)~\cite{heusel2017gans} evaluates individual frame quality by comparing feature distributions from a pre-trained model. Fréchet Video Distance (FVD)~\cite{unterthiner2019fvd} quantifies the distributional distance between real and generated videos, providing an overall assessment of video fidelity. Sync-C and Sync-D~\cite{chung2017out} evaluate lip synchronization from content and dynamic perspectives, with higher Sync-C and lower Sync-D scores indicating superior alignment with audio. Structural Similarity Index (SSIM)~\cite{wang2004image} measures structural consistency between ground truth and generated videos, while E-FID~\cite{deng2019accurate} provides a refined image fidelity evaluation based on Inception network features.

\noindent\textbf{Implementations.}
The experiments are conducted on a computing platform equipped with 8 NVIDIA V100 GPUs. Training is performed in three stages, with each stage consisting of 30,000 iterations and a batch size of 4. Video data is processed at a resolution of \(512 \times 512\). 
The learning rate is fixed at \(1 \times 10^{-5}\) across all stages, and the AdamW optimizer is employed to stabilize training. Each training clip comprised 16 video frames.
In the archived-clip motion-prior module, we set $\alpha=16$, $m=256$, and $n=16$.
In the present-clip motion-prior diffusion model, the number of layers \(L\) is set to 8, and the weighting factor \(\alpha\) in Eq.~\ref{alpha} is configured to 0.1 to balance the influence of prior motion information. This setup is chosen to optimize long-term identity preservation and enhance motion consistency within generated TalkingFace videos.

\subsection{Main Results}
We compare our method with several SOTA methods, including Audio2Head~\cite{wang2021audio2head}, V-Express~\cite{wang2024v}, AniPortrait~\cite{wei2024aniportrait}, SadTalker~\cite{zhang2023sadtalker}, Hallo~\cite{xu2024hallo}, EchoMimic~\cite{chen2024echomimic}, and {MegActor-}$\Sigma$~\cite{yang2024megactor}, from quantitative, qualitative, and user study.
\textbf{\underline{Unless otherwise specified,}} all methods do not use landmarks to ensure a fair comparison.

\begin{table}[t]
\centering
    \footnotesize
    \setlength{\tabcolsep}{0.8pt}
\begin{tabular}{l|cccccc} 
\toprule
\cellcolor{mygray}\textbf{Method} & \cellcolor{mygray}{\textbf{FID$\downarrow$}} & \cellcolor{mygray}{\textbf{FVD$\downarrow$}}  & \cellcolor{mygray}{\textbf{Sync-C$\uparrow$}} & \cellcolor{mygray}{\textbf{Sync-D$\downarrow$}}& \cellcolor{mygray}{\textbf{SSIM$\uparrow$}} & \cellcolor{mygray}{\textbf{E-FID$\downarrow$}} \\
\midrule
Audio2Head & 87.21 &1836.25 & 2.32 &13.92 &0.613 &3.12 \\
V-Express & 62.18 &1324.57 &5.45 &9.04  &0.674 &2.81\\
AniPortrait &56.11 &954.91 &6.37 &8.29 &0.706 &2.60 \\
SadTalker & 52.77 & 847.20 & 6.94 & 7.92 &0.724 &2.49 \\
Hallo & 51.35 &792.38 &6.85 &7.65 &0.728 &2.35 \\
EchoMimic & 49.20 & 751.44 & 7.06 & 7.18 & 0.737 & 2.31 \\
{MegActor-}$\Sigma$ & 48.57 & 724.40 & 7.22 &7.14 &0.745 & 2.29 \\
 \textbf{MCDM (Ours)}  & \textbf{42.08} & \textbf{656.71}& \textbf{7.84}& \textbf{6.69}& \textbf{0.779}& \textbf{1.97} \\
\bottomrule
\end{tabular}
\vspace{-0.3cm}
\caption{\textbf{Quantitative comparisons on TalkingFace-Wild.} MCDM achieves a significant advantage over other methods.}\label{tab:wild}
\vspace{-0.4cm}
\end{table}

\begin{figure*}[t]
 \centering
\includegraphics[width=0.95\linewidth]{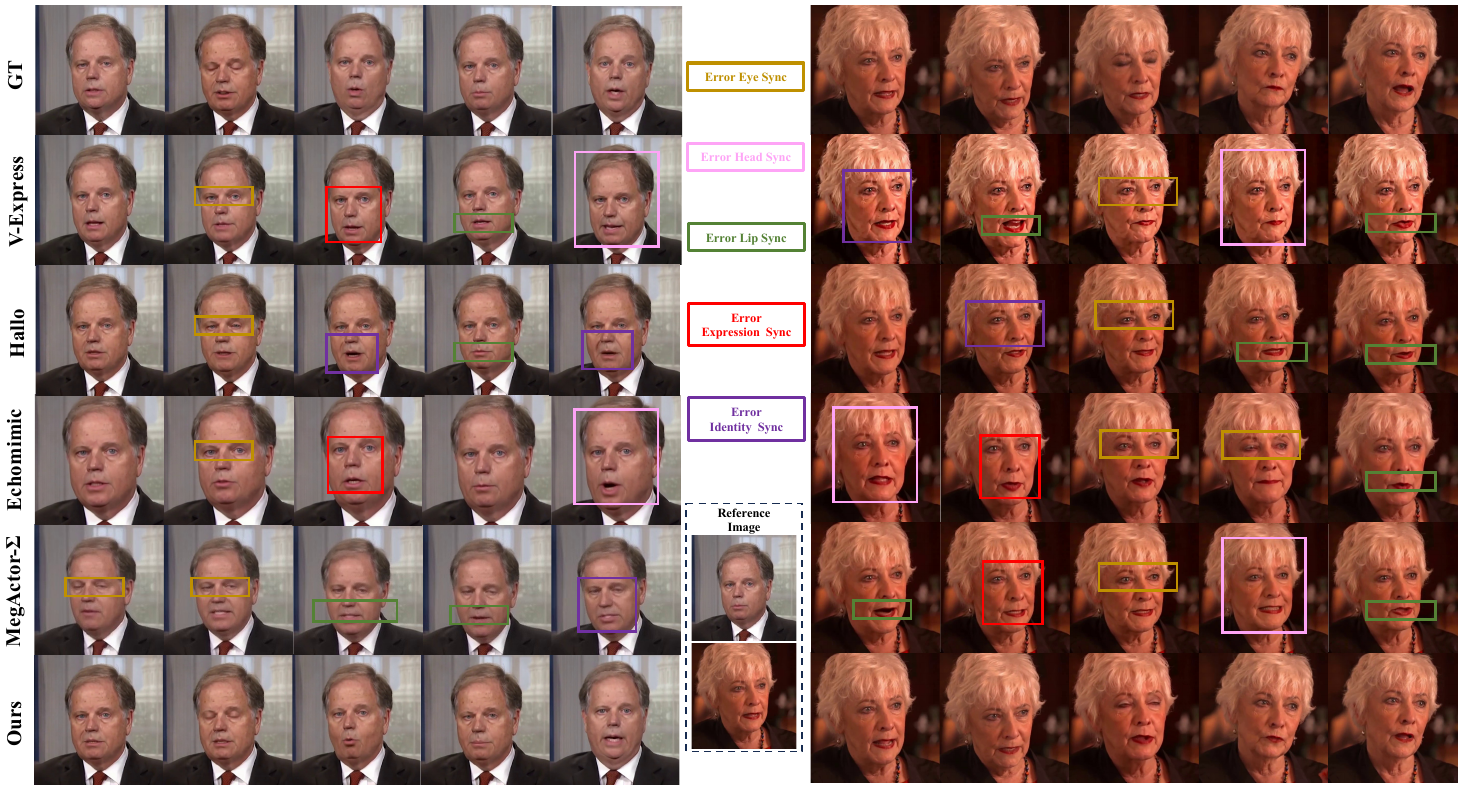}
\vspace{-.5cm}
     \caption{\textbf{Qualitative comparison on HDTF and CelebV-HQ.} Our method achieves the best generation results, particularly in identity consistency and motion detail.
     }
          \vspace{-.2cm}
     \label{fig:sota}
 \end{figure*}

\begin{figure*}[t]
 \centering
\includegraphics[width=0.9\linewidth]{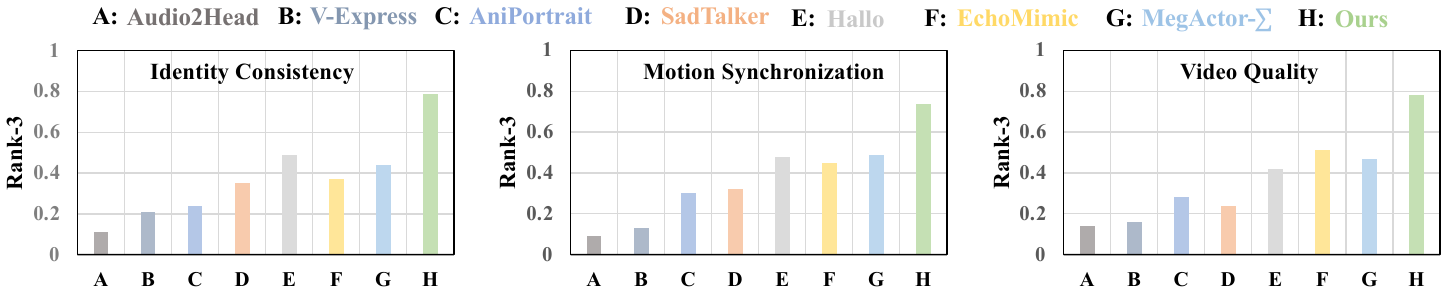}
     \vspace{-.5cm}
     \caption{ \textbf{ User study results} of identity consistency, motion synchronization, and video quality. Higher values indicate better performance.
     }
\vspace{-.3cm}
     \label{fig:user}
 \end{figure*}

\noindent\textbf{Quantitative Evaluation.}
Table~\ref{tab:sota} presents a quantitative comparison on the HDTF~\cite{zhang2021flow} and CelebV-HQ~\cite{zhu2022celebv}, illustrating the overall superior performance of diffusion-based methods compared to GAN-based methods. Our proposed MCDM achieves the best scores across all metrics, outperforming existing diffusion-based approaches. Specifically, MCDM achieves superior lip-sync accuracy, reflected in higher Sync-C and lower Sync-D scores, outperforming methods like EchoMimic~\cite{chen2024echomimic} and {MegActor-}$\Sigma$~\cite{yang2024megactor}, which show notable declines in synchronization quality. MCDM’s outstanding SSIM and E-FID scores also highlight its ability to generate visually appealing, temporally consistent content with precise lip synchronization.

Table~\ref{tab:wild} summarizes the quantitative performance on the proposed TalkingFace-Wild dataset. Consistent with results on HDTF~\cite{zhang2021flow} and CelebV-HQ~\cite{zhu2022celebv}, MCDM surpasses all competing SOTA methods across evaluation metrics, demonstrating marked improvements in visual quality and temporal consistency. Achieving the best FID, FVD, and an E-FID of 1.97, MCDM shows strong capability in generating high-fidelity TalkingFace videos under diverse conditions, effectively maintaining temporal coherence across audio, expressions, and lip synchronization.

\noindent\textbf{Qualitative Evaluation.}
Figure~\ref{fig:sota} provides a qualitative comparison of our method against other SOTA approaches. Compared to V-Express~\cite{wang2024v} and EchoMimic~\cite{chen2024echomimic}, our approach shows superior head and lip synchronization, benefiting from the audio-visual consistency introduced by motion priors. 
Additionally, unlike Hallo~\cite{xu2024hallo} and {MegActor-}$\Sigma$~\cite{yang2024megactor}, Our method accurately captures subtle facial actions, including blinks and expression nuances through the archived-clip, while better preserving identity consistency.
Overall, our approach demonstrates the best visual results.

\noindent\textbf{User Study.}
The quantitative and qualitative comparisons underscore the substantial advantages of our proposed MCDM in generating consistent TalkingFace videos. To further evaluate video quality, we conducte a user study, focusing on identity consistency, motion synchronization, and overall video quality. We randomly selected 10 cases, shuffled the generated videos from each method, and recruited 20 participants (10 male, 10 female) to provide rank-3 preferences. From Figure~\ref{fig:user}, our method consistently achieved the highest scores across all metrics in the user preference evaluation. This user study highlights the significant advantage of our approach in user-centric TalkingFace generation.

\subsection{Ablation Results}
We conduct an ablation study to assess the impact of each component in our method. Table~\ref{tab:ablation} shows the results: {w/o $F_a$} omits historical frame information, {w/o $F_{pc}$} adds an audio attention module for audio feature input, and {w/o $MTA$} applies a standard temporal attention module.

\begin{table}[t]
\centering
    \footnotesize
    \setlength{\tabcolsep}{2 pt}
\begin{tabular}{l|cccccc} 
\toprule
\cellcolor{mygray}\textbf{Method} & \cellcolor{mygray}{\textbf{FID$\downarrow$}} & \cellcolor{mygray}{\textbf{FVD$\downarrow$}}  & \cellcolor{mygray}{\textbf{Sync-C$\uparrow$}} & \cellcolor{mygray}{\textbf{Sync-D$\downarrow$}}& \cellcolor{mygray}{\textbf{SSIM$\uparrow$}} & \cellcolor{mygray}{\textbf{E-FID$\downarrow$}} \\
\midrule
{w/o $F_a$}  & 46.25  & 708.93 & 7.37  &7.05 & 0.749 & 2.25   \\
{w/o $F_{pc}$}  & 45.63 & 684.20 & 7.49  &6.97 &0.758 & 2.13  \\
{w/o $MTA$}  & 44.27 & 671.05 & 7.62  &6.84 & 0.771  & 2.04 \\
 \textbf{Ours}& \textbf{42.08} &\textbf{656.71} & \textbf{7.84}& \textbf{6.69}& \textbf{0.779} & \textbf{1.97}  \\
\bottomrule
\end{tabular}
\vspace{-0.3cm}
\caption{\textbf{Ablation results} on the TalkingFace-Wild dataset.}\label{tab:ablation}
\vspace{-.5cm}
\end{table}

\begin{figure}[t]
\includegraphics[width=0.95\linewidth]{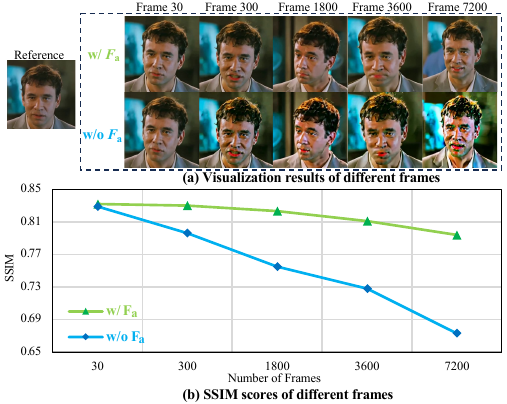}
  \vspace{-0.3cm}
\caption{\textbf{Visualization results and SSIM scores} during long-term generation. We find that w/ $F_a$ offers a distinct advantage in maintaining both identity and contextual consistency.
     }
 \vspace{-0.3cm}
     \label{fig:long}
 \end{figure}

\begin{figure}[t]
\includegraphics[width=\linewidth]{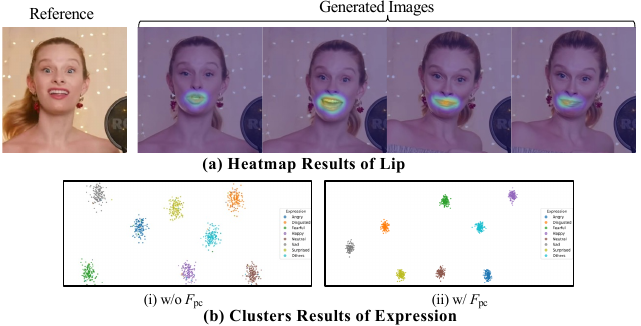}
    \vspace{-0.3cm}
     \caption{\textbf{Lip heatmap and expression cluster.}
We find that w/ $F_p$ effectively tracks the lip region and conveys expressions.
     }
 \vspace{-0.5cm}
     \label{fig:vis}
 \end{figure}

\noindent\textbf{Archived-Clip Motion-Prior.}
The results in Table~\ref{tab:ablation} show that removing historical frame information (w/o $F_a$) significantly degrades performance across all metrics, underscoring the importance of the archived-clip motion-prior. To further assess the effect of $F_a$ on long-term generation, we visualized frames 30, 300, 1800, 3600, and 7200 with corresponding SSIM scores, as shown in Figure~\ref{fig:long}. Figure~\ref{fig:long}(a) indicates that without the archived-clip (w/o $F_a$), identity consistency worsens with frame progression, resulting in visible artifacts and inconsistencies in head, mouth, and expression. In Figure~\ref{fig:long}(b), the SSIM scores highlight error accumulation increases with frame count, showing a rapid decline in (w/o $F_a$), while (w/ $F_a$) remains stable at a higher value. These findings validate the effectiveness of the archived-clip motion-prior in preserving both identity and temporal coherence over extended sequences.

\noindent\textbf{Present-Clip Motion-Prior.}
Similarly, excluding the present-clip motion-prior and injecting audio information directly via audio attention (w/o $F_{pc}$) leads to a drop in performance across all metrics. 
This decline highlights the effectiveness of the present-clip motion-prior in leveraging multimodal causality and temporal interactions to decouple and predict motion states, including expressions, lip movement, and head motion (see Table~\ref{tab:ablation}). 
To further validate this decoupling capability, we visualize heatmaps of the predicted lip tokens, as shown in Figure~\ref{fig:vis}(a), where the present-clip motion-prior accurately localizes and tracks lip motion. For expression decoupling, t-SNE~\cite{van2008visualizing} visualization of expression tokens reveals tighter clustering within each of the eight distinct emotion categories when using the present-clip motion-prior, indicating improved separation of emotional content from audio input.

\begin{figure}[t]
\centering
\includegraphics[width=0.9\linewidth]{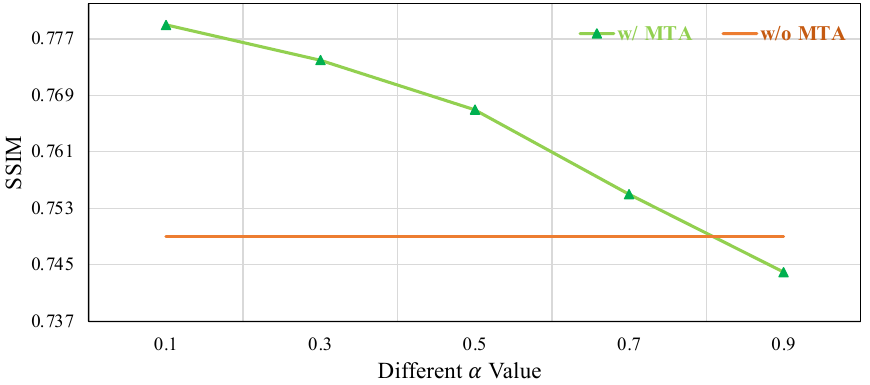}
   \vspace{-0.3cm}
     \caption{\textbf{SSIM results for different $\alpha$ values.}
Our method performs comparably well when the $\alpha$ value is smaller than 0.9.
     }
 \vspace{-0.3cm}
     \label{fig:alpha}
 \end{figure}

\noindent\textbf{Memory-Efficient Temporal Attention.}
Following the standard approach~\cite{hu2024animate}, we replace the proposed memory-efficient temporal attention with conventional temporal attention by directly summing $F_{ac}^{ref}$ and $F_{pc}^{ref}$. As shown in Table~\ref{tab:ablation}, this modification significantly degrades performance across all metrics. This drop in quality is primarily due to the absence of an update mechanism, which introduces gaps between the archived clip and the present clip, compromising video smoothness.
Next, we analyzed the effect of different $\alpha$ values in Eq.~\ref{alpha}, which control the update rate, on the model's SSIM performance, as shown in Figure~\ref{fig:alpha}. We observed that as $\alpha$ increases, SSIM gradually declines. When $\alpha$ is below 0.9, our approach significantly outperforms the w/o $WTA$  configuration. However, at $\alpha = 0.9$, the performance is weaker than w/o $WTA$, due to the excessive accumulation of historical frame information and a reduced proportion of the present clip. Consequently, we set $\alpha = 0.1$ as the default value in this paper.

\begin{table}[t]
\centering
    \footnotesize
    \setlength{\tabcolsep}{1.0pt}
\begin{tabular}{l|cccccc} 
\toprule
\cellcolor{mygray}\textbf{Method} & \cellcolor{mygray}{\textbf{FID$\downarrow$}} & \cellcolor{mygray}{\textbf{FVD$\downarrow$}}  & \cellcolor{mygray}{\textbf{Sync-C$\uparrow$}} & \cellcolor{mygray}{\textbf{Sync-D$\downarrow$}}& \cellcolor{mygray}{\textbf{SSIM$\uparrow$}} & \cellcolor{mygray}{\textbf{E-FID$\downarrow$}} \\
\midrule
 {B1}  & 42.49 & 668.24 &7.69  &6.78 &0.771 & 2.02  \\
{B2}  & 47.12  &721.17 & 7.30  &6.84 & 0.732 & 2.29   \\
 \textbf{Ours}& \textbf{42.08} &\textbf{656.71} & \textbf{7.84}& \textbf{6.69}& \textbf{0.779} & \textbf{1.97}  \\
\bottomrule
\end{tabular}
\vspace{-0.3cm}
\caption{\textbf{More results} of variant MCDM.}\label{tab:more_results}
\vspace{-.5cm}
\end{table}

\noindent\textbf{More Results.}
Table~\ref{tab:more_results} evaluates different design variants. In B1, Q-Former~\cite{li2023blip} replaces frame-aligned attention, while in B2, Reference UNet~\cite{hu2024animate} substitutes VAE with Reference UNet, omitting archived-clip information.
Results show that frame-aligned attention outperforms Q-Former by effectively capturing temporal context and integrating long-term dependencies. Additionally, using a frozen VAE with a trainable patchify layer proves to be an efficient alternative to the conventional Reference UNet.

\vspace{-0.2cm}
\section{Conclusion}
We presented the \textbf{M}otion-priors \textbf{C}onditional \textbf{D}iffusion \textbf{M}odel (\textbf{MCDM}) to address the challenges of long-term TalkingFace generation by achieving robust identity consistency and motion continuity. MCDM integrates three key innovations: an archived-clip motion-prior to enhance identity representation, a present-clip motion-prior diffusion model for accurate motion prediction, and a memory-efficient temporal attention to mitigate error accumulation over extended sequences. Additionally, we introduced the \textbf{TalkingFace-Wild} dataset, offering over 200 hours of multilingual video data across diverse scenarios. Experimental results demonstrate the effectiveness of MCDM, setting new benchmarks in long-term TalkingFace generation.

\section*{Impact Statement}
This paper presents the MCDM model, designed to enhance identity and temporal consistency in long-term TalkingFace generation. While MCDM contributes to the advancement of generative modeling, we recognize the potential ethical concerns, including the risks of misuse for creating deceptive content or spreading misinformation. We emphasize the importance of transparency in AI development and support the integration of detection frameworks to mitigate these risks. In alignment with ongoing efforts in responsible AI, we aim to ensure that the benefits of our work are balanced with its ethical implications, promoting safe and constructive applications in society.

\nocite{langley00}

\bibliography{example_paper}
\bibliographystyle{icml2025}


\end{document}